\documentclass{article}
\usepackage{PRIMEarxiv}
\usepackage[numbers]{natbib}

\usepackage[utf8]{inputenc} % allow utf-8 input
\usepackage[T1]{fontenc}    % use 8-bit T1 fonts
\usepackage{hyperref}       % hyperlinks
\usepackage{url}            % simple URL typesetting
\usepackage{booktabs}       % professional-quality tables
\usepackage{amsfonts}       % blackboard math symbols
\usepackage{nicefrac}       % compact symbols for 1/2, etc.
\usepackage{microtype}      % microtypography
\usepackage{lipsum}

\usepackage{amsmath}
\usepackage[utf8]{inputenc}
\usepackage[english]{babel}
\usepackage[dvipsnames]{xcolor}
\usepackage[ruled,vlined,noresetcount]{algorithm2e}
\usepackage{pdfpages}
\usepackage{graphicx}
\usepackage{caption}
\usepackage{subcaption}
\usepackage{amsfonts}
\usepackage{amsthm}

\usepackage{float}
\graphicspath{{figures/}{tables/}}
\usepackage{multirow}

\newcommand{\x}{\mathbf{x}}
\definecolor{beaublue}{rgb}{0.74, 0.83, 0.9}
\newcommand{\R}{\mathbb R}
\newcommand{\Lcal}{\mathcal L}
\usepackage{comment} 

\title{Feature Network Methods in Machine Learning and Applications}

\author{
  Xinying Mu, Mark Kon \\
  Department of Mathematics and Statistics\\
  Boston University, Boston, MA\\
  \texttt{\{xmu,mkon\}@bu.edu} 
}
\begin{document}
\maketitle

\begin{abstract}

A machine learning (ML) feature network is a graph that connects ML features in learning tasks based on their similarity. This network representation allows us to view feature vectors as functions on the network. By leveraging function operations from Fourier analysis and from functional analysis, one can easily generate new and novel features, making use of the graph structure imposed on the feature vectors.
Such network structures have previously been studied implicitly in image processing and computational biology. We thus describe feature networks as graph structures imposed on feature vectors, and provide applications in machine learning. 
One application involves graph-based generalizations of convolutional neural networks, involving structured deep learning with hierarchical representations of features that have varying depth or complexity. This extends also to learning algorithms that are able to generate useful new multilevel features.  Additionally, we discuss the use of feature networks to engineer new features, 
which can enhance 
expressiveness of the model.
We give a specific example of a deep tree-structured feature network, where hierarchical connections are formed through feature clustering and feed-forward learning. This results in low learning complexity and computational efficiency.
Unlike "standard" neural features which are limited to modulated (thresholded) linear combinations of adjacent ones, feature networks offer more general feedforward dependencies among features. For example, radial basis functions or graph structure-based dependencies between features can be utilized.
\end{abstract}

\keywords{machine learning \and feature networks \and hierarchical networks \and graph clustering }

\section{Introduction}

\subsection{Feature networks and functions on feature networks}
Over the last two decades the analysis of graphs through machine learning has gained considerable interest. This is partially due to the expressive capability of graphs to represent diverse systems across social sciences, natural sciences, and artificial intelligence. 
The concept of a feature vector is a cornerstone in machine learning, with numerous associated structures. In this paper, we propose a new approach that involves application of network/graph structures to feature vectors, involving the notion of `feature networks'. The formation of networks from data components is already very common \citep{yu2018learning, zhang2019acfnet}.  Nevertheless, we believe that this general network approach for organizing machine learning feature vectors is new, and that feature networks constitute a valuable extension to existing and newer machine learning methods.
This paper describes the principles and scope of this approach and describes several applications.

In supervised learning, a sample or object is typically represented as a feature vector. A dataset $\mathcal T ={(\x_i,y_i)}_{i=1}^N$ with feature vectors $\x_i=(x_{i1},\ldots,x_{ip})$ (comprising $p$ features $x_{ij}$) can be represented as a data matrix ${\bf X}$ with rows $\x_i$ of sample feature vectors and columns of features.  Intrinsically, graph or network connections, representing similarity indicators, can be established either among rows (samples) or columns (features).

In line with this, the graph structures for associated machine learning (ML) tasks may naturally involve two kinds of nodes: 1. In the context of \textbf{sample networks}, nodes symbolize training samples labeled $\mathbf{x}_i$ or $i$ in $\mathcal T$ (the rows of $\mathbf{X}$). 2. 
In \textbf{feature networks}, nodes correspond to features $j$ (the columns or column labels of $\mathbf{X}$).

We focus here on feature networks, and their usefulness in structuring machine learning feature vectors. Such networks have nodes (or vertices) defined as  individual machine learning features $x_i$ or their indices  $i$. The graph or network connections will be formed by prior or posterior feature relations, i.e., based on prior knowledge or on correlations derived from the training data. This will also extend to the generation of new features (nodes) from existing ones, for instance, new features formed as cluster averages or combinations of current features within feed-forward network structures.
Structuring feature vectors into these networks can enhance machine learning effectiveness using existing features and facilitate the development of new features from the existing ones.

{\bf From neural networks to feature networks.} In the recent advancements of deep neural networks, a key emerging attribute is their ability to `uncover' novel and significant features as single or multiple neuron outputs deep within the network. These are higher-order or derived characteristics of input feature vectors. Such features may encapsulate neuron activations sensitive to specific object classes (e.g., trees) at higher levels of a deep neural net, which we will refer to as {\it deep features}.

It becomes apparent that such extended deep features within a neural net, brought about by interactions of simpler features (encoded in lower-level neurons), are not exclusively attributed to neural nets and their specific neural activation rules. Rather, these features reside in neural signatures themselves as emergent features that could possibly be obtained, perhaps more readily, through more universal feature interaction rules. Thus, it seems logical to shift the focus away from specific neural architectures towards the understanding that emergent features can interact more broadly to generate novel (e.g., deeper) features.

Feature networks can be viewed as an extension of neural networks, where the primary entities becoming features encoded in network nodes. The weights of the network edges can range from real-valued feature correlations to Boolean ${0,1}$ connection values, and can also include more complex functional dependences of higher-level features on lower-level ones.

Network connections between data components are a well-established concept in literature \citep{min2009deep,tang2013deep,kim2013deep}. In this paper, we restrict these notions to the context of connecting features within ML feature vectors.
This concept has a number of prior origins, including the adaptation of neural networks into radial basis function (RBF) networks, as proposed by Girosi \citep{girosi1995regularization}. Included among examples of such network structures, Min et al. \citep{min2009deep} designed a scalable feature mapping method based on a pre-trained deep neural network, aiming to enhance the performance of $k$ nearest neighbor classifiers.
Tang (2013) \citep{tang2013deep} demonstrated successful learning by substituting the conventional softmax neural connection with a linear support vector machine. The learning connections subsequently aimed to minimize loss based on margin loss rather than cross-entropy loss. 
Similarly, Kim et al. \citep{kim2013deep} proposed a deep structure model using iterated regression capable of generating classification-important features, boasting commendable generalization capabilities.

Among the many other instances of graphical structures on data, these references include tree structures on features (i.e., with features as nodes), modeled in feed-forward neural networks with shallow or deep architectures. Network structures involving computed features can also incorporate general functional dependencies among linked features, e.g., including functional dependences modeled using RBFs.

When viewed as a feature network, a feedforward neural network has a graph structure where edges, formed as weights from lower to higher layers, define new features (neural activations) from previous ones. These weights encode the propagation of feature information, with each node representing a potentially significant feature, given the appropriate network configuration.

In a feature network where features are defined as functions of other features in adjacent nodes, the propagation of features can be encoded by simple 0-1 weights, if suitable. Here, a weight of 1 denotes a connection between two features, enabling them to exchange information. Specifically, a feature value $x_i$ can be determined from connected features $x_j$, along with a function $f_i$ such that $x_i=f_i({x_j:j \text{is connected to} , i})$.

In a fixed model, some reasonable constraints on the space of functions $f_i$ would be necessary, possibly based on computability. However, these could range from feedforward functions implemented in machine learning, such as support vector machine (SVM) output functions, to radial basis functions (RBFs) \citep{girosi1995regularization}. The feature network emphasizes features rather than the nodes computing them. This focus is conducive to network-based feature discovery, generating novel features from existing ones through extended network structures. A comprehensive ML feature network could consist of feature nodes forming primitive and higher-level features from an initial basic feature universe (for example, individual pixel activations in an image field). Distant nodes might contain highly abstracted versions of basic features in the initial network.

It's important to note that if the features of a machine learning algorithm are arranged into a network structure, then the positions of the features in the feature vector are what constitute the network connections. In this context, the feature values in a particular feature vector (data point) can be interpreted as a function on this network or graph.

Representing feature vectors as functions on graphs enables the application of several theoretical toolkits designed for such graph functions. The burgeoning field of Graph Signal Processing (GSP) \citep{shuman2012emerging, ortega2018graph} offers a fresh perspective for machine learning tasks, as it applies analogs of Fourier analysis tools to graph functions. GSP considers graphs as generalized metric spaces, wherein functions can be decomposed and analyzed using techniques borrowed from standard harmonic analysis on metrizable spaces.

In a broader context, many useful techniques from classical functional analysis can also be extended to such settings, thus enhancing the range of possibilities in machine learning applications, which we will explore further. Variations of this approach for different types of graphs have been investigated by other researchers \citep{dong2016learning,wu2020comprehensive,cheung2020graph}. However, to the best of our knowledge, the use of the feature network approach as a more general tool for machine learning has not been explicitly discussed previously.

\subsection{Deep learning with feature networks - an example in image processing}
Image recognition and processing provide a key example of how network structures can be applied to machine learning features. In this context, features are pixel intensities, which have a locality structure that can be represented as a graph (with adjacent pixels connected). This graph, seen as a feature network on pixel intensities, can be analyzed using the concepts discussed earlier.

Image processing algorithms often implicitly utilize an adjacency-based network structure of pixels for various tasks, such as denoising via smoothing or enhancing contrast. They  have also used more advanced methods from statistics and machine learning \citep{geman1993stochastic,takeda2007kernel,portilla2003image, li2009support, chambolle1998nonlinear}. Local Markov random field models \citep{li2012markov, geman1993stochastic} have been applied to image restoration, within statistical image denoising. Applications of kernel regression and smoothing in image processing and reconstruction have also previously gained attention \citep{takeda2007kernel,portilla2003image}, with more advanced methods such as support vector regression \citep{li2009support}.  These methods use natural adjacency structures on pixels, replacing them with network adjacency structures on corresponding features.

Pixels in images have Euclidean positions that naturally generate adjacency networks. For machine learning (ML) tasks in image recognition, such adjacency structures (which inform pixel positions) are implicitly used in tasks including unsupervised image improvement as well as supervised image learning.  In non-image machine learning contexts, feature vector indices no longer represent Euclidean positions, but  can still have natural graphical (or 'nearness') structures connecting different features. Such nearness structures represent prior or additional knowledge. For example, in image recognition, the prior knowledge is the known relative positions of pixels. In non-image tasks, the above prior knowledge from pixel position can be replaced by known or derived edge relationships among features, used similarly to improve machine learning.  Examples of such graphical relations include edges formed as known correlation or mutual information structures among features \citep{rapaport2007classification, mu2016differentiation}.

\subsection{Deep features:  Generalizations of deep and convolutional neural network structures}

There is a natural graph adjacency structure connecting pixel intensity features in image recognition.  In order to parallel the success of neural networks  for image recognition, we will also use deep layered graph structures.  Much as deeper and more complex features are encoded in neural intensities  in higher layers of a deep visual network, analogous complex and deep features can appear in general feature networks as well.

Deep learning neural networks can be seen as tools for extracting complex features from simple input feature vectors. Activations deep within the network are key outputs of the process. In feature networks a significant focus will be on formalizing such extraction of complex features and developing additional tools for them.

{\bf Convolutional structures on feature networks.}  
As mentioned, one basic illustration of feature network theory is the generalization of pixel adjacency graphs useful in image processing to more diverse network structures for input layers. This can generalize convolutional structures in neural networks \citep{bronstein2017geometric}. A convolutional neural network (CNN), widely used for image processing tasks \citep{lecun1989backpropagation,lecun1995convolutional}, can be seen as an example of a feature network.

Let's say our input feature vector comes from a grayscale image where the pixel intensities (the components of our feature vector) form a two-dimensional rectangular grid structure. If we create a correlation-based graph that connects pixels that have similar intensity patterns, we quickly get a graph that links neighboring pixels, both vertically and horizontally. This essentially reproduces the original geometric structure of the image.
Though this adjacency structure is what allows convolutional networks to process images effectively, it's worth noting that the structure can also be generalized and applied to a broad range of machine learning tasks that may not involve image processing.

When an image is fed into a convolutional neural network (CNN), the network first picks out small, local details. As the image moves through the network, these details get combined to form larger, more complex features. In essence, the CNN is able to extract and learn from features at multiple scales. This feature-extraction process has led to major advances in machine learning, particularly in deep learning and its many applications
\citep{lecun2015deep}. Broader analysis of CNNs and related graph characteristics have led to some keys of CNNs: local connection, shared weights among different local connections, and multiple layers. These approaches can also be important in solving problems in the graph domain, since: 1) graphs are more general locally connected structures, 2) shared weights (e.g. weight computations shared across different local connections on the graph) can reduce computational costs, and 3) multi-layer structures form a key to hierarchical pattern recognition for identifying features of various extents and abstraction levels.

With these shared characteristics, the graph neural network (GNN) has been a natural generalization of a CNN \citep{zhou2018graph}.
The metric structure of a convolutional network layer allowing convolutional windows is replaced by a looser graph similarity structure, again laterally within the same layer.
Because of their high performance and interpretability, GNNs have been broadly applied in graph analysis, \citep{henaff2015deep,kipf2016semi,monti2017geometric,zhou2018graph},
with convolutional structure techniques extended from visual deep learning.  Corresponding convolutional structures are also naturally framed in the context of feature networks as well as neural networks.

{\bf Preview of further applications: convolutional structures.} By analogy we will seek graph window partitions (analogous to convolutional image windows) that unify closely related features. Learning algorithm parameters identified for one such convolutional window can then transfer to other windows, hopefully reducing substantially learning times as compared with fully connected neural nets. 
The graph convolutional neighborhoods can be generated efficiently and form the receptive fields of a convolutional architecture, allowing the framework to learn effective graph representations.  Moreover the window clustering can be iterated until optimal performance achieved. 
Some simple examples of such convolutional graph structures on feature vectors are given in (Section 4).

{\bf Pooling in feature networks.}  Convolutional structures aggregate information between successive layers in neural networks as well as in feature networks.
The simplest form of such convolutional aggregation is recursive averaging over clusters within layers in a neural network, a process known as pooling.  Such pooling can also be useful in feature networks.
We will present an example that includes deep feed-forward hierarchical tree structures based on the underlying graph structure of feature vectors obtained through recursive feature clustering. Various feature mappings from one layer to the next can be perceived as a deep learning framework applicable to all classification algorithms.

Effectively, this is a feature network analogue of a neural network, with pooling occurring from one layer to the next. This gathering of related features into summary features at the next feature layer occurs either by external or intrinsic network relationships. Section \ref{pooling} provides an example of such a pooling feed-forward feature network.

{\bf Computational biology applications.} 
In order to summarize some additional applications of feature networks, it is worthwhile here to describe some prior work in computational biology that is related to the present discussion.

In computational biology, gene expression profiles are common in disease diagnostics and prognostics \citep{bellazzi2007towards,ramaswamy2003molecular}. Gene expression profiles are very noisy, and denoising methods have drawn attention for a number of years \citep{aris2004noise,tu2002quantitative}. Researchers have focused on using underlying structures within the space of genes and their expression values. 

Local averaging techniques have been used to denoise gene expression signatures \citep{fan2010regularization,Fan2010}.
Averaging of gene expressions among nearby genes forms denoised features (reinforcing similarities while cancelling individual variability from expression noise). Viewing a gene network as a base space with gene expression as a function on it is implicitly used by Rapaport, et al. \citep{rapaport2007classification}, who adapted classical signal processing techniques such as the discrete Fourier transform and spectral analysis \citep{rabiner1975theory} to denoise expressions by removing "high frequency" (noise) expression components on the gene network. 

Various underlying structures can be expressed in gene networks, which are feature networks with nodes consisting of gene expression features.
Among others, one way to obtain such a network structure is to connect genes in the same biochemical pathway with edges.
Such a network structure has been used to average gene expressions of `connected' genes, i.e., those in the same pathway, on the premise that connected genes will be co-expressed and thus have similar expression values.  The expressions of connected genes are then averaged in order to eliminate noise.
Thus by averaging expression values of genes in common pathways, we can filter out noise and get a clearer signal. This is analogous to the technique of pooling in neural networks.
 
Thus perhaps the simplest form of structuring gene expression signals is locally constant regularization, which averages gene expressions with those of other nearby genes. For example, \citep{kim2012pathway,su2009accurate,bild2006oncogenic,lee2008inferring,Wang2005gene} grouped genes into KEGG biochemical pathways, averaging expressions within pathways, on the premise that genes linked by metabolic steps should have similar differential expression values.  
In addition, searches of protein-protein interaction (PPI) networks (interpreted as gene-gene interactions) for highly interactive groups also yield such networks \citep{chuang2007network}. 

{\bf Smoothing features and regularization.}
We now briefly mention two additional applications mentioned at the end of this paper.  The first addresses improvements to standard machine learning classification via extraction of new and potentially very useful features that can be used with any ML algorithm.  Such features essentially exploit sets of features that are traditionally considered redundant because they are highly correlated.  This method parses the feature network into highly correlated cliques. Then, through graph signal processing (GSP) techniques (i.e. graph Fourier analysis via the Laplacian) `differentiate' redundant signals within cliques to offer very different information from different locations \textit{within} such cliques.  This is analogous to the contrast enhancement obtained from differentiating functions on the real line (Section \ref{smoothness}).

The second application uses analogous ideas for `contrast enhancing' machine learning feature vectors.  This is done in ways that extend the same process in visual images to provide new and additional information (Section \ref{regularization}).

{\bf Purpose.} 
The main goal of this paper is to explain the idea of feature networks and show how they can be applied in machine learning. We also aim to integrate ideas from graph signal processing and graph functional analysis, treating machine learning feature vectors as functions defined on a graph. This provides a new perspective on feature vectors and allows us to use methods from Fourier analysis and functional analysis that would not normally apply. The examples we'll give will demonstrate this new perspective.

\section{Construction of a feature network}
\subsection{Network components}
In machine learning, for feature vectors $\x = f = (f_1,\ldots,f_p)$ with labels $y$, consider a training set ${\cal T}=\{\x_i,y_i\}_{i=1}^n$ with data matrix 
$$
X=\begin{bmatrix}
\x_1^t\\
\x_2^t\\
\x_3^t\\
\vdots\\
\x_n^t
\end{bmatrix}
= \{X_{ij}\}_{ij},
$$
so that $X\in\mathbb{R}^{n\times p}$.  Here $\x^t$ denotes the transpose of $\x$. 
The \textit{feature network} for this problem is represented by a graph $V$ whose $p$ nodes are the features ${f_1,\ldots,f_p}$, or equivalently their corresponding indices ${1,\ldots,p}$. The edges connecting these features are assigned weights $w_{ij}\ \ (1\leq i,j\leq p)$ which define `connections', between features $f_i$ and $f_j$.  Such connections can reflect prior knowledge about the connected features (e.g. their correlation level in past datasets or known databases), or even empirical correlation within the given training set $\cal T$.  Let $W=(w_{ij})$ denote the matrix with entries $w_{ij}$.  The graph $G(V,W)$ is denoted as the ML problem's {\it feature network}.

 The graph $G$ on the feature set $\{f_q\}_{q=1}^p$ is assigned edge weights $w_{ij}$ that reflect, for instance:

\begin{enumerate}
    \item Empirical feature correlations observed in current or prior datasets. For example, in image recognition, correlations of pixel intensities (for a fixed set of registered pixels) usually reveal the geometrical structure of images (i.e. which pixel is adjacent to which). More generally even negative feature correlations (i.e., negative graph weights) can be used in certain graph algorithms, including  graph clustering algorithms. 
    \item Similarities based on mutual distances. Physical distances separating features $f_i$ and $f_j$ might have a form $$d(i,j) = \|f_i-f_j\|,$$
    which may represent an empirical average distance between the features. Such a distance can then be used to build a positive definite Gaussian kernel
    $$w_{ij} = \exp\bigg\{-\frac{d(i,j)^2}{\sigma^2}\bigg\}.$$
\end{enumerate}

\subsection{Deep hierarchical feature networks} \label{Sec1_6}

In image processing, convolutional neural networks reduce the cardinality of pixel-level feature vectors by pooling them into convolutional windows and subsequently subsampling these windows into deeper network layers. This convolutional pooling leverages the inherent multiscale clustering of the pixel grid.
On the pixel grid, such clustering respects the intrinsic (2D geometrical) metric and translational structure.

We can extend this concept of an image-based pixel adjacency network to a more general weighted, undirected feature network, denoted as $G = (V,W)$, where $V = (1,2,\ldots,p)$, and develop a similar hierarchical structure using multiscale clustering.  Let $G_0 = G$ be the graph structure (e.g., based on adjacency or correlation) for the feature input layer (set of input features). 

The aforementioned pooling strategy used for pixel features in convolutional windows can be generalized in a general feature network to involve local pooling of closely connected features. Features derived from images, such as pixel intensities, exhibit high correlations when originating from adjacent pixels. Consequently, a suitable extension of a local convolutional window within a feature network could be a cluster of features connected by strong mutual correlations.
  
Extending the analogy of pooling correlated machine learning features to the pooling of pixels in deep network convolutional windows, this process can be iterative. Similar to how a convolutional neural network's deeper layers encapsulate more extensive windows of visual pixel features, a convolutional feature network iteratively applies the correlational pooling process to higher feature layers, summarizing larger correlated subclasses of the initial set of features $f_1, \ldots, f_p$.
 
As in deep convolutional neural networks, this process is recursive, with features at each layer aggregating from feature clusters in the previous layer. Similarly, network weights in each layer aggregate from weights within clusters in the previous layer.  
For the $k$th layer in the deep structure with $G_k = (V_k,W_k)$, we let $C_k =  (C_{k,1},\cdots,C_{k,d_k})$  (with $C_{k,j} \subset V_k$) be a partition of $V_k$ into $d_k$ graph clusters based on $G_k=(W_k, V_k)$, where 
$V_k = \bigcup_{j=1}^{d_k}C_{k,j}$. 
With hard clustering algorithms, the clusters $C_{k,j}$ are mutually exclusive. With soft clustering, the clusters can overlap, i.e., each node can belong to more than one cluster. 

With the graph partition $C_k$, we can form a new network $G_{k+1} = (V_{k+1},W_{k+1})$ for the next layer, where 
$|V_{k+1}|=d_k$ is the cardinality of $V_{k+1}$, and the weights in $(k+1)$th layer are defined as
\begin{equation}
    W_{k+1,i,j} = \frac{1}{|C_{k,i}|\cdot|C_{k,j}|}\sum_{s\in C_{k,i}, t\in C_{k,j}} W_{k,s,t}, \mbox{\; if } i\neq j,
\end{equation}{}
i.e. an average of weights between the  clusters $C_{k,i}$ and $C_{k,j}$.

\begin{figure}[!ht]
\centering
\includegraphics[width=.8\textwidth]{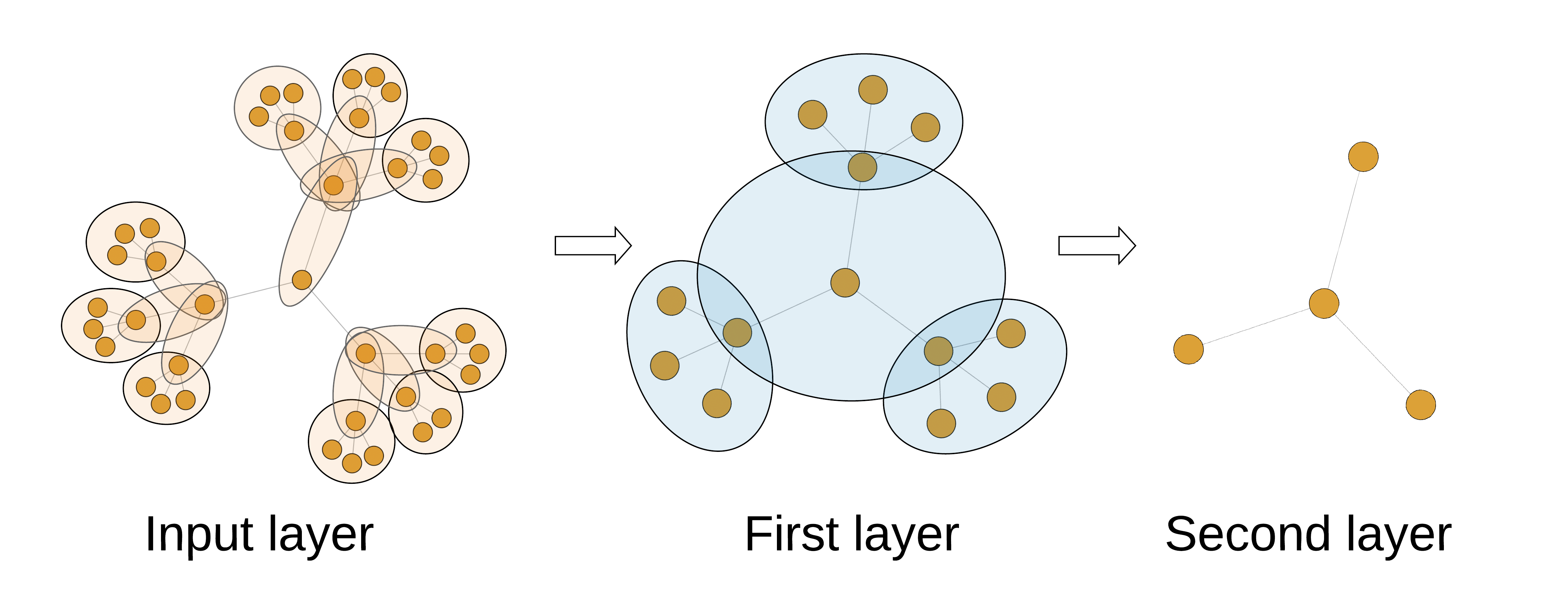}
\caption{Deep hierarchical structure for undirected Graph $G$ with two levels of clustering, i.e., two layers in the network learning.}
\label{fig:deep_structure}
\end{figure}

This architecture allows machines to feed-forward feature information in a structured way. Figure \ref{fig:deep_structure} below provides a simplified illustration of the deep hierarchical structure based on clustering.

\section{Tree Structured Networks}\label{network_methods}
In this section, we introduce several machine architectures that utilize different recursive convolutional definitions of aggregate features $f_k$ at nodes $k$ within the feature hierarchies, building upon the hierarchical convolutional structure described in Section~\ref{Sec1_6}.
We can define a general feed-forward feature structure as follows. If $f_{k} = (f_{k,1},\cdots,f_{k,d_{k-1}})$ is the input to layer $k$, let output $f_{k+1}$ be
\begin{equation}
    f_{k+1} = g_{k+1}\big(f_{k};\Theta,\{C_{k}\}),
\end{equation}
where $\Theta$ is a set of parameters, and $C_k$ is the collection of convolutional clusters at level $k$.

In this section, all computations have been performed using the statistical software package R, and accuracy represents balanced accuracy, i.e., arithmetic mean of sensitivity and specificity.

\subsection{Average pooling }\label{pooling}
Denoising of standard noisy functions on Euclidean spaces $\R^n$ typically assumes a function with largely independent additive noise (e.g. white noise), that can be mitigated through local averaging (e.g. blurring) of the data.  The effect is to diminish variance (by cancelling adjacent white noise components) while introducing minimal bias in the signal (if the underlying signal is smooth).  
Transferring these Euclidean denoising techniques to the denoising of structured feature vectors has been implicitly applied in various applications, especially in computational biology (e.g., \citep {rapaport2007classification}, \citep{fan2010regularization}). 

This convolutional approach averages adjacent features, usually based on a graph/network adjacency structure.
This concept can be extended to address denoising problems for feature vectors with underlying network or metric structures on the features (i.e., on their index space).  Note that in the specialized context of neural networks such averaging when done across a given layer is \textit{pooling}.
Numerous instances of this approach exist, including 
applications in computer networks \citep{ma2009identifying} and financial market analytics.  \citep{mandere2009financial,tumminello2007correlation}. Here we will primarily focus on biological examples.

Though we will consider the use of underlying network structures, in fact full distance-based structures, with features forming a metric space, can provide the most useful denoising. Clearly, any metric structure on a set of basic features can be interpreted as a network, with weights inversely defined by distances.

\textbf{Applications.}
For the datasets below we have used a training set to define a graph structure on the feature set, based on feature correlations.  These datasets involve gene expressions, where correlation structures give rise to gene coexpression networks.  We performed clustering using standard graph clustering algorithms to create convolutional windows formed by the resulting clusters.

\begin{figure}[!ht]
\centering
\begin{subfigure}{0.495\textwidth}
\includegraphics[width=\textwidth]{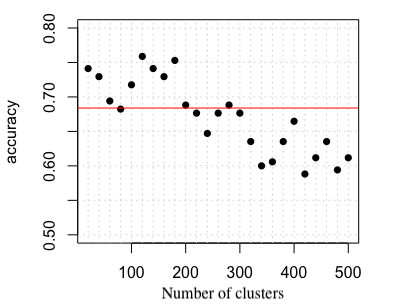}
\caption{CNS data}
\end{subfigure}
\begin{subfigure}{0.495\textwidth}
\includegraphics[width=\textwidth]{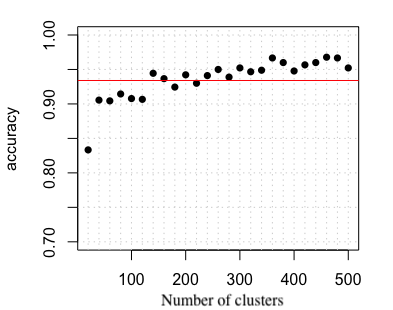}
\caption{Lung data}
\end{subfigure}

\caption{Classification performance with one layer average pooling for two cancer datasets. The red line in the graph indicates the benchmark performance by SVM with linear kernel.}
\label{P3_simu1}
\end{figure}

These convolutional clusters are pooled into simple summary features, namely cluster averages.
In this context, we limited our analysis to a single layer of convolutional pooling.  Fig \ref{P3_simu1} illustrates how the supervised classification performance is affected as the partition window size increases. The embryonal CNS Tumor dataset \citep{pomeroy2002prediction} includes 7129 genes and 34 patients in two classes (25 Classic, 9 Desmoplastic), while the Lung dataset \citep{TanGeman} contains 12533 genes and 181 patients with binary labels (150 ADCA, 31 Mesothelioma). 
We employed hierarchical clustering with Ward's method, varying the number of clusters for analysis.

The classification results are summarized in Fig \ref{P3_simu1}. Under a simplified model where a gene expression feature vector from a particular class $f=(f_1, \ldots, f_p)=g+\eta$ represents a sum of an underlying (average) signal $g=(g_1, \ldots, g_p)$ and an effectively independent noise (idiosyncratic) signal $\eta=(\eta_1, \ldots, \eta_p)$—which encompasses all effects beyond the averaged signal—the intrinsic noise levels in the CNS and Lung data are notably distinct.
In cases where feature vectors $f=g+\eta$ are dominated by noise $\eta$ over the signal, the relatively minor bias introduced by cluster averaging is overshadowed by a significant reduction in variance due to noise cancellation.  For such scenarios, forwarding these features through the feature network to the subsequent feature-averaged layer enhances the feature quality, thereby elevating cancer classification accuracy.
 What is indicated in the CNS dataset is that pooled cluster-averaged features diminish classification accuracy.  In other words, the loss of information from additional bias dominates the variance improvement from convolutional pooling. Consequently, this convolutional pooling results in a decline in classification performance as the pooled window (cluster) increases.  Conversely, this trend does not hold for the lung dataset.  Here, the decrease in variance
with increasing pooled window size enhances the feature vector signal, suggesting that noise dominates the second (Lung) dataset as opposed to the first (CNS) dataset.

This inference of substantial noise in the CNS data implies that a smaller number of pooled clusters yields smoothed features without inordinate information loss, ultimately enhancing classification performance. In contrast the Lung cancer gene expression vectors could be inferred to be less noisy, and over-smoothing (pooling windows that are too large) would decrease detail in feature vectors, jeopardizing performance. 
The interpretation in terms of intrinsic noise levels aligns with benchmark classification accuracies: Lung cancer achieves an accuracy of 94\% using a linear SVM, whereas CNS classification attains only 68\%.

We have additionally tested the same feature pooling approach on two well-known human breast cancer datasets \citep{vantVeer2002, Wang2005gene} with gene expression features. In both datasets gene expressions are used as predictors of breast cancer metastasis. Wang's breast cancer dataset includes 266 samples with 22,491 genes (92 samples with later 
metastasis, and 174 without metastasis). Van de Viever's breast cancer dataset includes 295 samples with 11,366 genes (101 with Metastasis, 194 without). For characterizing the underlying structure of the gene space, a gene network based on protein-protein interactions (PPI) is employed.
 
By applying graph convolutional windows and averaging features within clusters, feature vectors are again smoothed/denoised. 
These denoised expression feature vectors are then used for predicting breast cancer 
metastasis using a support vector machine (SVM) classifier. With minimal tuning of the regularization parameters, the area under the ROC curve is improved by 17 and 7 percentage points, respectively, over the SVM using raw gene expression features as input.
In this experiment, the pooling sizes for the three layers are set as $d_1 = 3,000$, $d_2 = 800$, and $d_3 = 100$ within the hierarchical structure. We also conducted a comparison between pooling achieved through hard clustering methods (e.g., spectral clustering) and soft clustering methods (e.g., fuzzy C-means \citep{pal1996sequential}). It's important to note that all accuracies provided in this section are balanced, meaning they have been adjusted to account for unbalanced test sets.

\begin{table}[!ht]
\centering
\begin{tabular}{c|c|cc|cc}
\hline
\multirow{2}{*}{input} & \multirow{2}{*}{features} & \multicolumn{2}{|c|}{Hard clustering} & \multicolumn{2}{|c}{Soft clustering} \\
  &  &Accuracy & AUROC & Accuracy & AUROC        \\
\hline
benchmark       &11356      &55.27 & 63.41& 55.27 & 63.41\\
1st layer       &2000		&65.19 & 75.28& 67.31& 78.15\\
2nd layer	    &500		&72.58 & 81.89&  75.44 & 79.21\\
3rd layer	    &100	    &57.78 & 68.22&  63.29 & 71.47\\
\hline
\end{tabular}

\caption{Average pooling with Wang breast cancer dataset. The input feature network is segmented as 2,000 subnetworks by clustering, and 2,000 new features are generated by averaging the subnetworks. The process continues recursively in a 2nd layer and a 3rd layer. Accuracy and AUROC are from an SVM model with a linear kernel.} 
\label{tab:p3_wang}
\end{table}

\begin{table}[!ht]
\centering
\begin{tabular}{c|c|cc|cc}
\hline
\multirow{2}{*}{input} & \multirow{2}{*}{features} & \multicolumn{2}{|c|}{Hard clustering} & \multicolumn{2}{|c}{Soft clustering} \\
 &  &Accuracy & AUROC & Accuracy & AUROC        \\
\hline
benchmark       &22233      &65.69 & 66.22 &65.69 & 66.22\\
1st layer       &3000		&73.32 & 75.28 &78.83 & 80.15\\
2nd layer	    &800		&71.21 & 74.89 &73.27 & 76.44\\
3rd layer	    &100	    &70.30 & 73.22 &72.01 & 75.29 \\
\hline
\end{tabular}
\caption{Average pooling with Van de Viever breast cancer dataset. The process is the same as Table \ref{tab:p3_wang} except for the sizes of clusters, and so numbers of features in each successive layer.} 
\label{tab:p3_van}
\end{table}
With the van de Viever breast cancer dataset (using the same outcome categories, based on cancer metastasis), the classification performance trend is similar. The accuracy is increased by $8\%$, and AUROC by 9\% with 3000 first layer features using hard clustering. With the same parameters and soft clustering, we have 13\% and 14\% improvements in accuracy and AUROC. Here, as the structure becomes deeper, performance becomes worse. This presumably means  features in the deeper layers are oversmoothed by averaging.  This is an example of the classic bias-variance problem, in which too deep a structure introduces excessive bias.

\subsection{Deep learning with SVM bagging}
The Support Vector Machine (SVM) is a widely used linear classification algorithm capable of achieving non-linear separation through appropriate kernel selection. It is worth noting that in a feedforward feature network, SVM continuous output scoring functions \citep{platt1999probabilistic} can generate useful feature nodes for the subsequent layer. 
This can be useful in in deeper network structures, since iteration of nonlinear maps provides the most useful growth in the scope of potential features at successive layers.

At its core, SVM is a linear discriminant classifier. When extended to support vector regression, the SVM classifier function $f(\mathbf{x})$ transforms into a regression function of the same form:

\begin{align}
f(\x)&={\bf \beta}\cdot \x + \beta_0 \nonumber \\
&=(\sum_i \alpha_i\x_i)\cdot \x + \beta_0 \nonumber \\
&=\sum_i \alpha_i(\x_i\cdot \x)+\beta_0\nonumber \\
&=\sum_i \alpha_iK(\x_i,\x)+\beta_0,\label{eq:9}
\end{align}
where $K(\x,{\bf y})$ denotes a linear kernel.

However, for our purposes, we are interested in feedforward feature networks that map feature values using nonlinear Radial Basis Function (RBF) feedforward functions similar to (\ref{eq:9}). It is known that combining RBF functions with varying parameters can approximate continuous and more general feedforward functions effectively \citep{girosi1995regularization}.  Thus 
general feedforward functions can effectively be attained using an appropriate (e.g. Gaussian) single RBF activation function in superposition as above.

The notion of a feedforward feature mapping using such kernel functions, some of them exhibiting strong discontinuities at each level, could promise some important universal properties in feature networks (e.g. including edge detection) that have been observed in natural and artificial neural networks.  Local filters in convolutional neural networks for edge detection have discontinuities built into them - these could be emulated by kernel functions $K$ with similar discontinuities.   In essence, a discontinuous kernel function can serve as an effective edge detector \citep{ranzato2007unsupervised, jarrett2009best}.

{\bf Algorithm details.} 
For this  application involving RBF, employing the same clustering strategy, we define nodes in the hidden layers using SVM scores translated into probabilities (see Platt, \citep{platt1999probabilistic}). For each local cluster $C_{k,j}$ in the $k^{th}$ layer, an SVM model is constructed using only features in the cluster, providing a predictive probability (using supervised learning) for the sample's class based only on the current cluster of features.
Define the $(k+1)$th layer output as the SVM probability score:
\begin{equation}
    f_{k+1,j} = Prob(\hat{y_j}| f_{k},C_{k,j})
\end{equation}
where $\hat{y_j}$ is the output of linear SVM classification for a subset of features $C_{k,j}$ based on the training set.  The final layer  consists of a single feature which combines through a final SVM the feedforward feature activations in the next-to-last layer.  

We note that nonlinearities are introduced in this feedforward feature network via the application of the Platt probability function to the linear outputs from the previous layer.

In contrast with this feature network, 
the networks of Sections \ref{pooling} and \ref{smoothness}, are formed through two constructions of node activations involving average pooling and a smoothness network. 

With SVM bagging as implemented in this approach, all nodes in the deep structure are derived from SVM models, thereby generating new features for deeper layers.
\begin{table}[!ht]
\centering
\begin{tabular}{c|c|cc|cc}
\hline
\multirow{2}{*}{input} & \multirow{2}{*}{features} & \multicolumn{2}{|c|}{Prostate1 } & \multicolumn{2}{|c}{Prostate2} \\
  &  &Accuracy & AUROC & Accuracy & AUROC        \\
\hline
benchmark       &12625      &89.53 & 93.52 & 75.03 & 83.82\\
1st layer       &1000		&90.69 & 96.01 & 82.91 & 82.61\\
2nd layer	    &100		&91.89 & 95.48 &  79.56 & 87.45\\
3rd layer	    &10	        &85.86 & 92.51 & 75.83 & 84.50\\
\hline
\hline
\multirow{2}{*}{input} & \multirow{2}{*}{features} & \multicolumn{2}{|c|}{Breast (wang) } & \multicolumn{2}{|c}{Breast (Van de Viever)} \\
  &  &Accuracy & AUROC & Accuracy & AUROC        \\
\hline
benchmark       &10000      &58.27 & 72.10 & 62.86 & 75.23\\
1st layer       &1000		&60.59 & 72.21 & 67.53 & 70.99\\
2nd layer	    &300		&61.77 & 72.65 & 65.22 & 68.89\\
3rd layer	    &80	        &62.84 & 73.03 & 65.92 & 68.07\\
\hline
\end{tabular}
\caption{SVM bagging network performance with four cancer datasets. The input feature network is segmented as 1000 and 2000 subnetworks by clustering for Prostate data and Breast data separately, and the new features are determined by linear SVM classifier from the features in subnetworks. The process continues for the 2nd layer and 3rd layer.  The final classification forms the output of the fourth layer.} 
\label{tab:p3_wang_svm}
\end{table}

\section {Graph Laplacian Methods}\label{laplacian_methods}

\subsection{Introduction to the smoothness discriminant}\label{smoothness}
The above view of an ML feature vector as a function on a feature network, akin to a pseudo-geometric or even geometric structure, can help to recruit pseudo-analytic and analytic techniques, particularly Fourier analysis, to ML.
In particular, this can bring some of the tools of graph signal processing to the preprocessing of ML feature vectors. It turns out that various Fourier-like and more generally functional analytic operations on feature vectors, expressed as functions in this framework, can  extract information that is otherwise obscured.  In particular operations graph differentiation and anti-differentiation through the application of the graph Laplacian $\mathcal L$ can serve as useful feature for discrimination tasks on ML feature vectors.  

In this context we consider pseudo-Sobolev space operations -- operations that would involve Sobolev spaces if they were performed on manifolds or Euclidean spaces -- involving powers of the Laplacian, as tools for enhanced ML discrimination.  This is rooted in an analogy of feature vectors (now \textit{feature functions}) to Sobolev (differentiable) functions on Euclidean spaces.  Such functions can in a sense be considered more `adapted' to Euclidean geometry if they have higher Sobolev smoothness.  In particular, a function $f(x)$ on $\R^p$ with a small Sobolev norm $||f(x)||_s$ exhibits `less' or slower variation  among adjacent (nearby) points.  By extension, a graph function with similarly small graph Sobolev norm $||f(x)||_s$ (defined below) will exhibit less variation among adjacent nodes, i.e., effectively respecting the adjacency structure of the underlying graph geometry.

In light of this, we can define an ML feature vector $f(x)$ to be better `adapted' to a given underlying feature network geometry $G$ if its Sobolev norm $||f(x)||_s$ is small for large $s>0$.  This would be equivalent to saying that feature vector $f$ belongs to the same class (i.e., the same training data) that generated the network $G$ if $||f||_s$ is small.

This norm serves as a measure the `adaptation' of single feature vectors to feature networks trained from a specific class of data.  Thus we define a feature function $f$ to be \textit{adapted} to the space $G$ on which it is defined when $f$ changes `slowly' relative to $G$.  For functions $f$ defined on networks $G$ this would translate to minimal variations in $f$ among nodes with strong connections, which will correspond to small graph Sobolev norm.  

In scenarios where a predefined feature network $G$ is generated from known feature correlations (e.g., gene coexpression) or other predetermined feature relations (e.g. protein interaction networks), feature vectors $f$ can be measured against the feature network $G$ based on their adaptation to $G$, measured by such graph Sobolev norms.  Just as Sobolev norms on $\R^p$ measure function smoothness (by viewing the Sobolev norm as a `smoothness penalty'), graph Sobolev norms of feature vectors $f$ within prior network structures $G$ measure their adaptation to $G$.  

We have developed a classifier based entirely on this measure of adaptation, using Sobolev norms
of feature vectors on a prior known or posterior derived feature network (e.g. feature correlation) structures.  

This classifier is based on a functional $\phi(f)$, defined below in (\ref{smoothness_penalty}) as a \textit{smoothness penalty} for a feature vector $f$.  Thus given two training classes $A$ and $B$ (e.g. different cancer subtype gene expression classes), new samples $f$ can be classified as A or B based on either prior (pre-training) or posterior (post-training) feature network structures $G_A$ and $G_B$ (e.g. gene coexpression networks) for cancers $A$ and $B$.  Given network structures $G_A$ and $G_B$, with corresponding Laplacian matrices ${\mathcal L}_A$ and ${\mathcal L}_B$, the ordered feature pair $\Phi(f)\equiv(f^T\Lcal_A f,f^T\Lcal_B f)$ will exhibit a low first penalty component if $f$ is adapted to (arises from) $A$, and a low second component if $f$ is adapted to $B$.

This new feature map $\Phi$ proves effective independently of standard features in cancer classification. In Table \ref{tab:smoothness_classfier} we provide comparative measures of the classification accuracy with the 2D feature space formed by this pair  using a linear discriminant, as opposed to standard machine learning classifiers in the original gene expression feature space.  The dataset here refers to that in section \ref{pooling} on prediction of metastasis in breast cancer - further details can be found there.

We remark here that a feature network $G$ based on correlations may involve negative edge weights.  Ideally this should not interfere with the effectiveness of using the Laplacian to compute their 'derivatives'.  However, one can show that for graphs with negative edge weights, the corresponding Laplacian $\Lcal$ is no longer positive definite.  Since positive definiteness is desirable for various applications, we use here a modification of the standard Laplacian operator, denoted as the \textit{positive Laplacian}.

For a graph $G = (V,W)$, the {\it modified smoothness penalty} for any function $x: V\rightarrow \mathbb{R}^p$ on $G$ is defined as:
\begin{equation}
\phi^*(x) = x^T\mathcal{L}^*x= \frac{1}{2}\sum_{W_{ij}\geq 0} |W_{ij}|(x_i-x_j)^2 + \frac{1}{2}\sum_{W_{ij}\leq 0} |W_{ij}|(x_i+x_j)^2,\label{smoothness_penalty}
\end{equation}
where $\mathcal{L}^* = D^* - W$ is the positive Laplacian matrix, and $D^*={\rm diag}(d_i^*)$ is a diagonal matrix with entries $d_{i}^* = \sum_{j}|W_{ij}|$.

Recall that the standard graph Laplacian has the form $\Lcal=D-W$, where $D={\rm diag} (d_i)$ is the \textit{degree matrix}, with $d_i=\sum_jW_{ij}$ is the sum of the weights in the $i^{th}$ row of the weight matrix $W$.  To form the \textit{positive Laplacian} $\Lcal^*$, we replace $D$ by $D^*= {\rm diag}\  d_{i}^*$, with $d_i^*=\sum_j|W_{ij}|$.  Thus $\Lcal^*=D^*-W$.  This positive Laplacian is what is used in place of $\Lcal$ in our simulations in the following paragraph.

We note that the classifier based on the two dimensional vector $\Phi(f)$ relies entirely on network-based features effectively independent of basic gene expression features used in benchmark SVM methods. This means that the feature map $\Phi$ is sufficiently complex that the resulting feature vector $\Phi(f)$ forms effectively a new (novel) feature vector. Notice that the smoothness features, used \textit{alone} in the last column of  Table \ref{tab:smoothness_classfier}, provide accuracies comparable to those achieved by benchmark SVM methods.

\begin{table}[!htb]
\centering
\begin{tabular}{c|c|c|c}
\hline
Data		& SVM(linear) & SVM(RBF) & Smoothness features \\
\hline
Van de Viever Breast Cancer Data	&70.63	&74.83	&77.13	\\
Wang Breast Cancer Data	&63.53	&65.79	&70.09 \\
Lung Cancer Data	&94.07	&95.14	&92.31	\\
\hline
\end{tabular}
\caption{Accuracy of classifiers for binary class datasets} 
\label{tab:smoothness_classfier}
\end{table}

\subsection{Multiple smoothness features via network clustering}\label{multiple_smoothness}

The notion of convolutional pooling based on feature networks, as described in Section \ref{Sec1_6} and illustrated in Section \ref{pooling} can also be extended through a modification of the above ML classifier in Section \ref{smoothness}.  
The features constructed in Section \ref{smoothness} measure the `smoothness' of (test) feature vectors viewed as functions on (trained) feature network graphs.  While the above method can train a feature network and test the smoothness of feature vectors as functions on this network, it is also possible to pool basic gene expression features $f_i$ into a convolutional structure, i.e. to cluster the features.  Each convolutional feature window (network cluster) would then adopt its own local classifier on test vectors $f^t$, based on their (local) smoothness within the respective feature clusters.

More generally, the method we describe above can be extended to compute individual smoothness features identical to those in section \ref{smoothness}, but based only on sub-graphs of the network.  Specifically, one can generate successive layers of a feature network in which each node at level $n$ of the network summarizes a single cluster at level $n-1$ by generating a smoothness feature based solely on this cluster.  Iteration of this process through layers of the network effectively forms a hierarchical clustering of the graph $G$, with each node summarizing the smoothness feature of its cluster in the previous layer.
These multiple cluster-level features are then integrated into a single feature vector used for classification.

Since feature vector smoothness relative to the feature network is an efficient feature map for classification \citep{mu2016differentiation}, such (local) smoothness of test feature vectors $f$ restricted to sub-graphs can form a useful collection of sub-sampled smoothness features, one from each convolutional window.

Consider a sub-graph $G_{k,j} = (C_{k,j},W_{k,j})$ consisting only of nodes in cluster $C_{k,j}$ (i.e., the $j$th cluster at clustering level $k$),  
where  $W_{k,j}$ represents the submatrix of the $k^{th}$ level weight matrix $W_k$ corresponding to the rows/columns in cluster $j$. 
Let $f_k$ be the vector of smoothness penalty features in the $k^{th}$ layer of this network. The $j^{th}$ node in the $(k+1)$th layer $f_{k+1,j}$ represents the smoothness penalty of the prior vector $f_k$ restricted to the corresponding subgraph in the $k^{th}$ layer, i.e.,
\begin{equation}
    f_{k+1,j} = \big[h(f_k,C_{k,j})\big]^T \mathcal{L}_{k,j} \big[h(f_k,C_{k,j})\big] = \sum_{s,t \in C_{k,j}} W_{k,st}(f_{k,s}-f_{k,t})^2
\end{equation}
where $h(f_k,C_{k,j}) = \{f_{k,t}\}_{t\in{C_{k,j}}}$ is a function that selects just features from cluster $C_{k,j}$ in layer $k$, 
and $\mathcal{L}_{k,j}$ represents the standard graph Laplacian of $G_{k,j}$.
A designated subset or a union of all of the smoothness features $f_{k,j}$ now form a multi-scale smoothness feature vector for classification.

\textbf{Applications.}
We utilized gene expression datasets described in \citep{TanGeman} to evaluate this approach. 
Data in all tables represent accuracy levels in percent. Table \ref{tab:p3_tan} displays the mean accuracy levels in percent over 10-fold cross-validation runs for all tested classifiers and all nine datasets.
In this scenario, only one smoothness layer beyond the initial feature layer is applied, and the number of clusters is determined by the Community Detection algorithm \citep{newman2006modularity} for clustering graphs. 
The number of clusters (chosen by the algorithm) ranged from 2 to 10. Note that here only smoothness features were used on these datasets in the last two columns of the table.

\begin{table}[!ht]
\centering
\begin{tabular}{c|cc|c|c|c|c}
\hline
Data	& No. samples	& No. features	& SVM(linear) & SVM(rbf) &Smoothness(\#clusters) & Smoothness (single)\\
\hline
 CNS       &34		&7129	&76.47 &82.35	&73.53 (3)	&94.12\\
Colon	    &62		&2000	&87.1  &74.19	&70.97 (3)& 72.58\\
Lung	    &181	&12533	&98.9	&98.9	&99.45 (5)& 97.79	\\
Leukemia	&72		&7129	&94.44	&91.67	&95.06 (4) & 97.22	\\ 
GCM		&280	&16063	&86.43	&85	&87.86 (7) & 80.71\\
DLBCL		&77	    &7129	&97.4	&92.21	&90.91 (3) & 89.61	\\ 
Prostate1	&102	&12625	&89.22	&88.24	&85.29 (4) & 85.29\\
Prostate2	&88		&12625	&75	&79.55	&80.41 (3) & 80.68\\
Prostate3	&33		&12625	&100	&100	&100 (2) & 100\\
\hline
\end{tabular}
\caption{Accuracy of classifiers for binary class expression datasets. The terms 'SVM(linear)' and 'SVM(rbf)' indicate SVM with linear kernel and radial basis function kernel as applied to the original data. The term 'Smoothness' indicates that the classification results come from the smoothness network features (subnetwork smoothness penalty) with linear SVM classifier, using either multiple clusters or the entire graph.} 
\label{tab:p3_tan}
\end{table}

To bolster the dataset size and more carefully examine the algorithm, we utilized simulated data (see below Table \ref{tab:p3_tan_simu2}). 
The number of features was reduced to $p=200$ through random sampling. For all cases, the simulation sample sizes were set to $n=300$ for each class. We generated simulated ('simulized') normal data from the real data above with \textit{equal} mean $0$ for the two classes. Moreover, the above two class covariance matrices $\Sigma_A$ and $\Sigma_B$ were adjusted to be more similar, making the two classes more difficult to discriminate.  This  led to two datasets $X^{[1]} \sim N(0,\Sigma_1)$, $X^{[2]} \sim N(0,\Sigma_2)$. Here, $\Sigma_1 = b\Sigma_A+(1-b)\Sigma_B$ and $\Sigma_2 = (1-b\Sigma_A)+b\Sigma_B$, where $\Sigma_A$ and $\Sigma_B$ are from the real data, and $b=0.3$ in this simulation.  Note these simulized data have identical means and more similar variances, compared with original data above, which explains their somewhat lower benchmarks.

\begin{table}[!ht]
\centering
\begin{tabular}{c|cc|c|c|c}
\hline
Data	& No. samples/class	& No. features	& SVM(rbf)  & sub-network Smoothness & single Smoothness\\
\hline
 CNS       &300		&200	& 75.80 & 84.28 & 81.93\\
Colon	   &300		&200	& 60.80 & 66.88 & 67.15\\
Lung	   &300		&200	& 69.07 & 82.52 & 82.63\\
Leukemia   &300		&200	& 67.60 & 81.18 & 79.85\\
GCM		   &300		&200	& 62.55 & 70.58 & 70.07\\
DLBCL		&300	&200	& 65.97 & 78.23 & 75.55\\
Prostate1	&300	&200	& 66.47 & 76.45 & 74.55\\
Prostate2	&300	&200	& 62.77 & 76.25 & 77.33\\
Prostate3	&300	&200	& 74.68 & 84.33 & 83.83\\
\hline
\end{tabular}
\caption{Accuracy of classifiers for binary class expression simulation (simulized) datasets. These are simulated datasets with covariance structure identical to corresponding real datasets.  The number of sub-networks is determined by community detection algorithm. Note this is discussed in Section \ref{multiple_smoothness}} 
\label{tab:p3_tan_simu2}
\end{table}

\subsection{Feature vector regularization}
\label{regularization}
Feature networks on feature vectors can also offer a means to regularize these vectors, i.e. denoise them to conform with prior information based on the feature network.  This concept has an analogy in image processing, where features (pixel intensities) are arranged in a grid-network based on pixel adjacency.  Images naturally respect this network structure, with adjacent pixels exhibiting highly correlated intensities.  Two means of exploiting this structure for image recognition involve contrast enhancement (sharpening of contrast between adjacent features) and denoising (blurring of adjacent features).

Analogously, given the availability of adjacency structures in general feature networks, both processes -- contrast enhancement and blurring -- can prove useful, depending on the level of 'blurring' (mixing of adjacent features) versus noise (additive independent feature values) that a given feature vector suffers from.  This is in analogy with the same dichotomy for image-based feature vectors.

{\bf Fourier regularizations used in image processing.}
In classical image recognition, image transformations implementing (a) blurring or, alternatively, (b) contrast enhancement, can be effective for obtaining images that are correspondingly (a) less noisy or (b) less blurred, respectively.  It is reasonable to expect that analogous 
processing could in some cases similarly improve the quality of (noisy) feature vectors from any dataset in which features form some principled network structure, in some sense parallel to the adjacency network structure in pixel values.  These two processes of  (a) image denoising and (b) deblurring are complementary here to the extent that the first diminishes and the second amplifies high frequency components of a signal. To the extent that we choose to view images as functions on graphs (where pixels are nodes), these two transformations generalize to two complementary processes on graph functions, and, by extension feature vectors (viewed as graph functions).  Image blurring ('softening') for noise reduction corresponds to convolution and local averaging, akin to application of negative powers of the Laplacian (or of a constant plus the Laplacian).  Equivalently, this corresponds to mapping the feature vector into a lower (i.e. smoother) Sobolev space. This essentially represents a form of anti-differentiation, or (fractional) 'integration' of the feature vector.

Conversely, contrast enhancement would thus correspond to the application of positive powers of the Laplacian, with the power scaled to the required contrast enhancement in the feature vector of interest.  This then corresponds to a (fractional) differentiation of the feature vector.  

{\bf Analogs in feature networks.}  While fractional integration (denoising) and fractional differentiation (deblurring) are well-established in image processing, a similar approach needs development for other feature vectors arising from datasets with natural feature network structures.  In particular, a `regularization parameter' reflecting an extent of (deblurring or denoising) regularization needs to be defined, represented by the power of the Laplacian (negative or positive, respectively) that optimizes recovery of the feature vector of interest.  Barring other methods for setting regularization parameters, cross-validation presents itself as a means to determine optimal feature vector regularizations.  

In the context of cross-validation, it remains to identify a metric for measuring the signal improvement in a feature vector from this type of pre-processing. One approach is to gauge 
`predictiveness' in a dataset i.e. the quality of cross-validation predictions based on such pre-processing.  For discrimination within datasets between two classes $A$ and $B$, this metric could translate to cross-validation-based predictive accuracy.  This would be based on a training/test set whose feature vectors have been pre-processed using the above scale of regularizations, ranging from negative (softening) through positive (contrasting) powers of the graph Laplacian.  

This approach has been applied here to benchmark cancer datasets from \citep{TanGeman}.  We applied this Laplacian transformation $\Psi(X;s) = \mathcal{L}^s X$ to each subject's gene expression feature vector $X$, now represented as a feature function on the (gene co-expression) feature network.  We proceeded to compare the classification performance using subjects' original, untransformed feature vectors $X$ as benchmarks. In the table, Benchmark columns are based on using the original feature vectors $X$, while the 'New features' column represents classification performance based on the transformed (smoothed or sharpened data) $\{\Psi_A(X;s),\Psi_B(X;s)\}$. Here $\Psi_A(X;s)$ denotes the regularization transformation based on the class A network and its Laplacian, and similarly for $\Psi_B(X;s)$.  Combining these two transformations simultaneously doubled the number of features in the feature vectors. The performance measures give accuracy using an SVM machine with an RBF kernel. The smoothing/sharpening parameter $s$ was determined by a cross-validation grid search.  The Laplacian used was the positive Laplacian $\Lcal=\Lcal^*$, defined in Section \ref{smoothness}.

\begin{table}[!ht]
\centering
\begin{tabular}{c|cc|cc}
\hline
Data	&   No. of genes & No. of samples & Benchmark &	New features\\
\hline
CNS	& 7129 & 34 & 82.94 &	85.29 \\
Colon &	2000 & 62 & 79.03&	85.81 \\
Lung& 12533 & 181 &	95.45&	99.45 \\
Leukemia& 7129 & 72 &	93.61 &	97.22\\
GCM& 16063 & 280 &	79.50 &	83.21\\
DLBCL& 7129 & 77 &	88.40 &	92.21 \\
Prostate1&	12600 & 102 & 94.12 &	92.16 \\
Prostate2& 12625 & 88 &	81.82 &	80.68\\  
Prostate3& 12626 & 33 &	100 &	100 \\
\hline
\end{tabular}
\caption{Data summary and classification results. Entries in the last two columns are percentage accuracies on corresponding datasets.}
\end{table}

\section{Discussion}

As discussed extensively above, an advantage of feature network structures is that feature vectors themselves can be represented differently, in that a real-valued feature vector $f = (f_1,\cdots,f_p)$ becomes a function $f:V\rightarrow\mathbb{R}$ on a graph/network whose vertices $V$ are features $f_q$, or more properly their indices $q\in \{1,\ldots,p\}$.
This function takes the values $f(q)=f_q$ for $q\in V$.
This perspective allows for \textit{generative} graph structures, where edges represent functional connections generating new features $f(q)$ from adjacent ones $\{f(r): r \ {\rm is\  a\ neighbor\ of\ } q\}$. 
Feature vectors, traditionally represented as real-valued vectors, can thus be seen as functions on graphs with values generated from adjacent ones via such connections. This enables feature engineering through network construction.

Above, graph nodes are feature vector indices, with functions on the graphs representing feature vector component values.
In machine learning applications we are concerned with the graph structures (feature connections), and functions on them, representing feature vectors.

Network structuring of features allows incorporation of prior (or posterior) information, e.g. feature correlations or mutual information, into machine learning (ML) feature vectors.  This can yield enhanced methods for supervised learning (Section \ref{smoothness}), as well as improved feature vectors based on prior knowledge (Section \ref{regularization}). 

{\bf Generalized connections.} 
Feature networks can unveil underlying structures of ML feature vectors, extending the capabilities of traditional  learning models. 
Replacing networks of neurons with networks of features can capture network learning more effectively. Indeed, deep neural networks produce their most significant value through their extraction of `deep features' represented by individual or groups of neurons.  The transition from neural networks to feature networks makes this value added an explicit one.
Lower level (more simple) fundamental units (neurons) are replaced by higher level ones (features) as atoms of 
computation, with a higher and more immediate added value, yielding networks with more compact and useful structures.

An analogy exists with continuous complexity theory \citep{blum1998complexity,blum2004computing,traub1988information,traub1994information}, which reformulates classical  (bit-level) computation and complexity theory in terms of higher level aggregate basic components.
The complexity of an algorithm (e.g. computational solution of a PDE) is then measured in terms of such coalesced operations, each counting as a single step in a higher level `continuous' algorithm. 

In a similar vein, more general feature variable values can replace neural activations as fundamental values on a network.
Connections between nodes need not be restricted to feedforward neural functions (e.g. a sigmoid composed with a linear combination) of adjacent features; feature values can be more general functions of neighbors, within a graph structure on ML features.

Connections and computed functions can be taken from an allowed collection of (possibly stochastic) functional relationships.  For example, one feature value $f_1$ may be connected to a group of feature values $\{f_i\}_i$ and computed as a nonlinear SVM classification or regression function of those features, joined nonlinearly at the receiving node as a kernel SVM target function.

{\bf Convolutional structures.}
In this paper (Section \ref{network_methods}, \ref{laplacian_methods}) we have obtained deep feature network structures by clustering, and demonstrated five applications with three different mechanisms for feature propagation among connected nodes (e.g. feed-forward feature propagation, including averaging and SVM feedforward).
In our pooling propagation mechanisms the middle layer has been defined via local average values, i.e. linear combinations of features. We have defined a quadratic form involving local smoothness penalties in a new layer reflecting smoothness properties of previous layers. All of these methods involve clustered convolutional structures in which local groups of features compute new single summary (convolutional) features in subsequent layers of a feature network.

Features propagating into deeper convolutional layers can also also arise via classification algorithms such as SVM (note these are not necessarily linear machines since different kernels can be in play at each layer).
All three methods of aggregating and propagating information from groups of feature nodes provide domains of improved classification.
Similarly, the idea of convolutional feature networks can be generalized to convolutional networks with backward propagation.

In standard convolutional neural nets, if $f_{k} = (f_{k,1},\cdots,
f_{k,d_{k-1}})^t$ is the input to layer $k$, the output $f_{k+1}$ into layer $k+1$ can be defined to be:
$$f_{k+1,j} = g\big(\sum_{i\in C_{k,j}} \omega_{k,i}f_{k,i}\big)$$
where $g$ is the activation function, $\omega_k$ represent training parameters, and $C_{k,j}$ represents the $j^{th}$ cluster in layer $k$.

Thus the concept of convolutional pooling structures, widely used in image processing feature vectors, can be extended to similar pooling in feature networks based on correlation-clustering windows.  Since adjacent pixels in image feature vectors are the most highly correlated, our method specializes to standard convolutional windowing algorithms when data are image-based.
More generally, convolutional feature networks can be represented as:
\begin{equation}
   f_{k+1} = g\big[(W_k\odot M_k)f_{k}\big]. 
\end{equation}
Here $M_k = [1_{C_{k,1}},\cdots,1_{C_{k,d_k}}]^T$ is the clustering mask with dimension of $d_{k+1}\times d_{k}$; the $i^{th}$ row $1_{C_{k,i}}^T$ has $0$'s in all positions except those of the $i^{th}$ cluster in layer $k$.
The matrix $W_k$ is the weight matrix with the same dimension as $M_k$. 
Here $\odot$ represents an element-wise product, so that the $i^{th}$ row of $W_k\odot M_k$ is the $i^{th}$ row of $W_k$ (connecting layer $k$ to layer $k+1$), masked so the only non-zero entries correspond to the $i^{th}$ cluster in the $k^{th}$ layer, via the mask $1_{C_{k,i}}$.
Thus only neurons in the $i^{th}$ cluster in layer $k$ are connected to the $i^{th}$ neuron in the $(k+1)^{th}$ layer. 
Here the function $g$ acts identically on each component of its argument.
The clustering mask $M_k$ guarantees that each node in the $(k+1)^{th}$ layer is restricted only to gaining information from a single cluster in the previous layer, as the backpropagation process continues. This is a generalization of backpropagation in CNN structures.

Notationally, we let $G_k$ denote the $k^{th}$ layer of the network $G$, with initial weights $W_k=B_k$ being feedforward from layer $k$ to layer $k+1$, and $C_k$ defining the clustering of layer $k$.  The backpropagation Algorithm 1 below evolves the initial weights $B_k$ iteratively into a sequence $W_k$. Below $X_k$ is the full feature vector representing the activations of neurons in the $k^{th}$ layer.  

\begin{algorithm*}[h]
\caption{Training of graph 
convolutional layers}
1. Given an initial deep hierarchical structure as a series of subgraphs $G_k = (V_k, W_k,C_k)$, $k = 0,1,\cdots,K$, and learning rate $\alpha$:\\
Forward Pass:\\
2. Input data: $X_k$\\
3. layer output: $X_{k+1}^t 
= g\big[(W_k\odot M_k)
X_{k}^t\big]$\\
4. Loss function: $J(\hat{y},y)$\\
Backward Pass:\\
5. Compute gradient w.r.t input: $\frac{\partial }{\partial X_{k+1}} J$\\
6. Compute gradient w.r.t weights: $\frac{\partial }{\partial W_k} J = \big(\frac{\partial }{\partial X_{k+1}} J \times g'\times X_k^t\big) \odot M_k$\\
7. Iterate weight matrix as: $W_k: = W_k + \alpha \frac{\partial }{\partial W_k} J$\\
\end{algorithm*}

This iterative algorithm generalizes the steps of CNN filtering and pooling within hierarchical network structures, reducing number of parameters at each stage.
In particular, note that the convolutional step above is entirely analogous to standard convolutional methods in image-based convolutional nets, with a geometric adjacency structure replaced by a correlation-based adjacency structure.

In real-world applications, network clustering can be computationally expensive due to the quadratic memory requirement ($O(N^2)$) and computational complexity ($O(\log(N^3))$) for a dataset with $N$ features \citep{newman2006modularity}.    
To address this, a pre-trained deep network structure, analogous to those in image processing/recognition, can be advantageous in many tasks, including gene-related ones (with gene expressions as feature vectors). Once deep structures are provided, the number of parameters and computation times can be significantly reduced, resulting in much smaller learning complexity compared to fully connected layers. 
With this deep hierarchical graphical clustering structure, graph-based feature maps can be applied to generate new features, potentially enhancing the performance of machine learning algorithms.

\section{Acknowledgements}  This work was partly supported by the NSF DMS (DMS 2319011) and the NIH NIGMS 
(NIH R01GM131409).

\bibliographystyle{plainnat}
\bibliography{references.bib}
\appendix

\end{document}